\documentclass[12pt]{article}
\usepackage{amsmath}
\usepackage{amsthm}
\usepackage{amssymb}
\usepackage{graphicx}
\usepackage{enumerate}
\usepackage{natbib}
\usepackage{url} % not crucial - just used below for the URL 
\usepackage{hyperref}
\usepackage[ruled,linesnumbered]{algorithm2e}
\newtheorem{Theo}{Theorem}

\newtheorem{Lem}{Lemma}

\usepackage{tikz}
\usetikzlibrary{shapes.multipart, positioning}

\newcommand{\blind}{1}

\begin{document}

\def\spacingset#1{\renewcommand{\baselinestretch}%
{#1}\small\normalsize} \spacingset{1}

%%%%%%%%%%%%%%%%%%%%%%%%%%%%%%%%%%%%%%%%%%%%%%%%%%%%%%%%%%%%%%%%%%%%%%%%%%%%%%

\if1\blind
{
  \title{\bf Efficient Human-in-the-Loop Active Learning: A Novel Framework for Data Labeling in AI Systems}
  \author{Yiran Huang\thanks{Equal contribution. The authors gratefully acknowledge \textit{the National Natural Science Foundation of China (Grant No.12271270). }We are grateful to the editor, an associate editor, and four reviewers for their insightful comments and constructive suggestions. The code is available at \texttt{https://github.com/Yiran-Huang/Active-Learning-with-Multiple-Questions}.},\hspace{.2cm}Jian-Feng Yang\footnotemark[1]\\
    School of Statistics and Data Science, LPMC $\&$ KLMDASR,\\Nankai University, Tianjin 300071, China\\
    and\\
    Haoda Fu\thanks{Corresponding author} \\
    University of North Carolina at Chapel Hill\\
    }
  \maketitle
} \fi

\if0\blind
{
  \bigskip
  \bigskip
  \bigskip
  \begin{center}
    {\LARGE\bf Efficient Human-in-the-Loop Active Learning: A Novel Framework for Data Labeling in AI Systems}
\end{center}
  \medskip
} \fi

\bigskip
\begin{abstract}
Modern AI systems rely heavily on labeled data, yet labeling is often expensive and labor-intensive, especially when requiring special skills such as reading radiology images by physicians. To most efficiently use experts' time for data labeling, one promising approach is human-in-the-loop active learning. However, traditional active learning methods are limited to single-label queries and fail to leverage more flexible query types in many real-world settings. In this work, we propose a novel active learning framework with significant potential for application in modern AI systems. This framework incorporates different query schemes and jointly determines which question to ask and which data points to query. We also introduce a method to integrate full and partial information obtained from these diverse queries. In addition, we develop an innovative model-agnostic exploration and exploitation framework to filter out redundant samples. Experiments on five datasets, including two real-world image datasets, demonstrate that the proposed framework substantially outperforms all other methods. These results highlight the framework's promise for applications across a range of scientific domains.

\end{abstract}

\noindent%
\vfill

\newpage
\spacingset{1.9} % DON'T change the spacing!
%-----------------------------------------------------------------
\section{Introduction}
\label{sec:intro}

A diverse range of AI systems has emerged for practical applications. Large language models, such as Generative Pre-trained Transformers (GPT) \citep{vaswani2017attention,radford2018improving}, have demonstrated exceptional capabilities in natural language processing tasks, including understanding, generating, and interpreting human language. Similarly, reinforcement learning models have been successfully applied to real-world scenarios, such as autonomous driving \citep{kiran2021deep}. These models are data-hungry, requiring a significant amount of labeled data for effective training \citep{zhao2023survey}. This poses challenges in data collection, large-scale annotation, and data quality improvement. Across various scientific domains such as language processing, reinforcement learning, medical imaging and speech recognition, there exists a substantial volume of unlabeled data and a relatively small amount of annotated data \citep{budd2021survey,zhu2022introduction}. It is prohibitive for traditional supervised model training, especially for deep learning models, because labeling a significant portion of data can be impractical. One solution to this dilemma is human-in-the-loop active learning algorithms. Active learning is a process in which humans interact with the learning system to improve its efficiency. It aims to identify the most informative unlabeled data and entrust them to oracle experts for labeling. The high quality of these selectively labeled data points significantly improves model performance. As a result, the overall labeling requirement can be reduced to a manageable level. Criteria from different perspectives are used to recognize informative data, including uncertainty-based methods \citep{settles2008analysis,nguyen2022measure}, Bayesian approaches \citep{houlsby2011bayesian,kirsch2019batchbald}, representative-based methods \citep{ash2019deep}, diversity-based methods \citep{zhdanov2019diverse}, density-based methods \citep{xu2007incorporating}, cluster-based methods \citep{wang2017active}, disagreement-based methods \citep{cohn1994improving,hanneke2011rates,houlsby2011bayesian,hanneke2019surrogate}, distance-based methods \citep{xing2002distance,yang2012bayesian,deng2023query}, assembling methods \citep{hsu2015active}, and data-driven approaches \ {\citep{konyushkova2017learning}}. In this paper, we introduce a novel human-in-the-loop active learning framework that allows not only standard label queries but also more flexible and efficient queries. We focus on uncertainty-based methods, as they are among the most widely adopted active learning strategies and can be applied across a wide range of models.

Beyond selecting the most informative samples, balancing the trade-off between exploration and exploitation is crucial, as this balance helps mitigate the ``cold start'' problem in active learning \citep{yuan2020cold,yi2022pt4al}. The exploration and exploitation allocate a proportion of the budget to gather information from areas with limited knowledge, because the model is likely to be inaccurate during the early stages of active learning and may misidentify the informativeness of each data point (see, for example, \cite{ren2021survey}). To address the need for balancing exploration and exploitation in active learning, we propose a novel adaptive framework. This mechanism improves query efficiency and helps the algorithm better utilize limited annotation budgets.

In this work, in addition to identifying the most informative data, we investigate different ways to annotate the data. Traditional active learning methods primarily focus on which data to query but often neglect how to label the data. Motivated by this limitation, we aim to develop a method that addresses both the challenges of data labeling and quality enhancement by considering various perspectives on labeling process. Specifically, traditional active learning algorithms query only one type of question: ``what is the class of this point?'' (referred to as a ``Class'' query) and receive a fully informative answer about the point. The differences among algorithms lie in their information measurements. In many practical scenarios, it is possible to query innovative questions such as ``are all of these data from a specific class?'' or ``is any of these data from a specific class?'' (referred to as ``All'' and ``Any'' queries, respectively). For example, in epidemiological testing, doctors conduct pathological tests to determine whether a patient has a disease and to identify the specific illness. By pooling samples from several patients and asking ``does any of these patients suffer from a specific illness?'', experimenters can obtain the result with a single laboratory test \citep{hogan2020sample,song2022review}.

Extending active learning to incorporate multiple question types involves several challenges. First, the new questions may yield partial information. It is crucial to effectively leverage this information to improve model fitting. Second, with an enriched set of available questions, we must not only select specific realizations but also decide which question to query. Third, as previously noted, balancing exploration and exploitation remains a significant challenge.

To address these challenges, we propose a new active learning framework whose novelty lies in the design for efficient human-in-the-loop learning. The paper first introduces a method to construct a probabilistic model using information gathered from various questions, including both full and partial information. Additionally, the paper proposes an uncertainty-based active learning framework using information gain. Unlike traditional methods that rely solely on full label queries, our framework introduces a flexible query system that supports a diverse set of questions, including full label queries and more efficient binary questions. These new questions not only reduce annotation cost, but also enhance efficiency by allowing information from multiple data points to be gathered simultaneously. The framework also automatically selects both the question and the data to be queried. In parallel, a data-driven exploration and exploitation framework is proposed. The framework can be embedded in a large number of active learning algorithms. The core idea of the framework is to automatically filter out data points likely to contain redundant information, based on their model-guided distances, before each active learning iteration. This distance metric is supervised and well-suited for high-dimensional data such as images. The exploration and exploitation framework is also leveraged to extend our method to a batch active learning setting, which is often much more efficient \citep{hoi2006batch,sener2017active,zhdanov2019diverse,ash2019deep}.

The benefits of querying these innovative questions are threefold. (1) The cost of posing such questions may be significantly lower than that of asking ``What is the class of the point?''.  In many areas, a human annotator can perform comparisons for multiple images in parallel, with annotation costs that are nearly constant or increase at a sublinear, sometimes logarithmic, rate with the number of images \citep{thornton2007parallel,buetti2016towards}. The structural difference between ``Class'' queries and ``All/Any'' queries mirrors the distinction between simple feature-based detection and complex object identification tasks \citep{wolfe1994guided}. For instance, consider the MNIST dataset. Answering a ``Class'' query requires determining the exact label among all possible categories. In contrast, answering an ``All'' query can be decomposed into two simpler operations: determining whether the images belong to the same class, and checking whether the shared class equals the queried digit. If the images are not from the same class, the annotator can immediately respond ``No''. Similarly, an ``Any'' query returns the answer ``Yes'' once a positive instance is found. By leveraging parallel perception, binary verification, and early stopping, ``All'' and ``Any'' queries can reduce annotation effort compared with repeated multi-class labeling. Moreover, in some cases, determining the exact label of a data point can be difficult, whereas determining whether it belongs to a specific class can be much easier. (2) Such queries can provide information about multiple points in a single iteration, though it may be partial information. (3) These queries can efficiently explore the sample space, as a single question may involve data from multiple regions.

The paper is organized as follows. Section \ref{sec:Background} provides background on the active learning problem and introduces key notations. Section \ref{sec:method} elaborates on the method for integrating both full and partial information, and introduces the active learning framework with multiple questions. The exploration and exploitation framework is also shown. Section \ref{sec:Theory} establishes a theoretical foundation supporting the proposed method. Section~\ref{sec:Simulation} illustrates the proposed active learning method through simulations on three datasets, and Section~\ref{sec: real data} further demonstrates its effectiveness using two real-world datasets. Section~\ref{sec:Concluding Remarks} concludes the paper and brings insights for future studies. All proofs are deferred to the supplementary material. The code is available at \href{https://github.com/Yiran-Huang/Active-Learning-with-Multiple-Questions}{our GitHub repository}.
%-----------------------------------------------------------------
\section{Background and Notation}
\label{sec:Background}
For convenience, all fraktur symbols in this paper denote sets. The full dataset is assumed to contain $N$ data points, where each $x_i \in \mathcal{X} \subset \mathbb{R}^d$ is associated with a label $y_i \in \mathcal{Y} = \{1, 2, \dots, C\}$; that is, $x_i$ belongs to class $y_i$, for $i = 1, 2, \dots, N$. The full data is denoted by $\mathcal{D}_{full}=\{(x_i,y_i)\}_{i=1}^N$. While most labels in $\mathcal{D}_{full}$ are unknown, we retain this notation for convenience. Following the standard assumption in active learning, let $\mathcal{D}_{0}$ represent a small fraction of $D_{full}$ as annotated data. We further denote by $\mathcal{D}_{full}^x=x_{1:N}$ all points without their labels from $D_{full}$, where $x_{1:N}=\{x_1.\dots,x_N\}$. $D_0^x$ is similarly defined. Though most labels in $\mathcal{D}_{full}$ are unknown, $\mathcal{D}_{full}^x$ is fully observed. Predicting the label $y$ for a given input $x$ requires constructing appropriate models. The paper focuses on the probabilistic model approach. Let $\mathbb{P} \subset \mathbb{R}^C$ denote the space of probability vectors; that is, for any $p \in \mathbb{P}$, $p_c \geq 0$ and $\sum_{c=1}^C p_c = 1$. The probabilistic model can be written as $p(\cdot;\theta):\mathbb{R}^d\to \mathbb{P}$, where the $c$th component, $p_c(x;\theta)$, represents the predicted probability of $x$ from class $c$. The model is trained using the cross-entropy loss, and the unknown parameter $\theta$ is estimated via
$$\hat\theta=\operatorname{argmin}_{\theta}\frac{1}{|\mathcal{D}_0|}\sum_{(x,y)\in\mathcal{D}_0}l(x,y;\theta) = \operatorname{argmin}_{\theta}\frac{1}{|\mathcal{D}_0|}\sum_{(x,y)\in\mathcal{D}_0}-\log p_y(x;\theta),$$
where $|\mathcal{D}_0|$ represents the size of $\mathcal{D}_0$. For notational simplicity, we write $\hat{\theta} = \theta(\mathcal{D}_0)$ to indicate that $\theta$ is trained on $\mathcal{D}_0$. Typically, the performance of the model built by the initial training set is unsatisfactory, thereby motivating the use of active learning. Traditional active learning methods only query one question, ``what is the class of $x$?". The paper considers an innovative scenario where multiple questions are available to be inquired, allowing data to be labeled from diverse perspectives.

We consistently use the uppercase letter $Q$ to denote the general form of how a question is queried, while the lowercase letter $q$ refers to a specific realization. Suppose there are $ \tilde k +1$ different questions in total, denoted by $Q_0,\dots,Q_{ \tilde k }$. Let $\mathcal{A}_k$ denote the set of all possible answers for question $Q_k$, where $k \in \{0, \dots, \tilde{k}\}$. For example, consider a question $Q_1$ of the form ``are $x_1,x_2$ from the class $c$?''. Given two specific points $x_{i_1},x_{i_2}$ and class $c = 1$, if we query ``are points $x_{i_1},x_{i_2}$ from class $1$?'' and the feedback is ``Yes'', the gathered information can be represented as $q = \{(x_{i_1},x_{i_2} ), c = 1\}$ and $a = 1$. 

Suppose question $Q_k$ has been queried $n_k$ times, with each realization and its corresponding answer denoted by $q_{ki}$ and $a_{ki}$, respectively, for $i = 1, \dots, n_k$. We denote by $\mathcal{D}_k=\{(q_{ki},a_{ki})\}_{i=1}^{n_k}$ the information obtained from question $Q_k$, and define $\mathcal{D}=\bigcup_{k=0}^{ \tilde k }\mathcal{D}_k$ as the total information gathered. Without loss of generality, we designate the question ``What is the class of $x$?'' as question 0, whose corresponding dataset $\mathcal{D}_0$ coincides with the training set introduced earlier. Throughout the paper, we focus on three types of questions ``what is the class of $x$?'', ``are all of $x_{1:m}$ from class $c$?'', and ``is any of $x_{1:m}$ from class $c$?'', which we abbreviate as ``Class'', ``All'', and ``Any'', respectively. By varying the value of $m$, multiple questions (more than three) can be generated.

%-----------------------------------------------------------------
\section{Methodology}
\label{sec:method}
Probability serves as the natural bridge linking different questions to the true labels. When class probabilities are known or can be predicted, the probabilities of answers to any question can likewise be computed or estimated. Define $\Pr(a|q;\theta)$ to be the predicted probability of answer $a$ given realization $q$ and parameter $\theta$.
\subsection{Loss functions for questions}
\label{subsec: loss f}
To utilize the information of each question $Q_k$, we define the loss function
$$l_k(q,a;\theta) = -\log \Pr(a|q;\theta).$$
The parameter $\theta(\mathcal{D})$ is then obtained by minimizing
$$\frac{1}{|\mathcal{D}|}\sum_{k=0}^{ \tilde k }\sum_{(q,a)\in\mathcal{D}_k}l_k(q,a;\theta).$$
The cross-entropy loss across different questions is a natural choice, as it establishes a more coherent relationship between the various questions and class probabilities compared to alternative loss functions. Specifically, if we inquire ``are all of $x_{i_1},\dots,x_{i_m}$ from class $c$?" and the answer is ``Yes'', then the loss for that realization and answer is
$$-\log\left\{\Pr(a_{ki}=1|q_{ki};\theta)\right\}=-\sum_{j=1}^m \log\{p_c(x_{i_j};\theta)\},$$
which equals the sum of losses for data $(x_{i_j},y_{i_j})$ for $j=1,\dots,m$. If we query ``is any of $x_{i_1},\dots,x_{i_m}$ from class $c$?" and the feedback is ``No'', then the loss becomes
$$-\log\{\Pr(a_{ki}=1|q_{ki};\theta)\}=-\sum_{j=1}^m \log[1-p_c(x_{i_j};\theta)],$$
which is equivalent to the sum of losses for the individual questions ``is $x_{i_j}$ from class $c$?" with the answer ``No'' for all $j=1,\dots,m$.

\subsection{Active learning method}
\label{subsec:AL}
The key to active learning is to estimate the amount of information that can be gained from querying a question. For example, commonly used criteria such as entropy or (predictive) variance estimate the information gain from querying ``what is the class of $x$?'' as 
$$H(x;\theta) := -\sum_{c=1}^C p_c(x;\theta)\log p_c(x;\theta) \quad\text{and}\quad V(x;\theta) := \sum_{c=1}^C p_c(x;\theta)\left(1- p_c(x;\theta)\right).$$
In this subsection, we first define a simple form of the information gain function, and then gradually extend it to a more sophisticated form suitable for our multi-question setting. Define a naive form of information gain function $G(\cdot||\cdot):\mathbb{P}\times \mathcal{Y}\to \mathbb{R}$. Then the expected information gain of question ``Class'' can be expressed as
$$\text{Gain}(x;\text{Class},\theta) := E_{p(x;\theta)}\left[G(p(x;\theta)||a)\right] = \sum_{c=1}^C p_c(x;\theta)G(p(x;\theta)||a=c).$$
One can verify that $G(p(x;\theta)||a=c)=-\log p(x;\theta)$ and $G(p(x;\theta)||a=c)=1- p(x;\theta)$ correspond to the entropy and variance criteria, respectively.

To generalize the information gain, we first slightly modify the definition of $G$. Now, suppose $G:\mathbb{P}\times\mathbb{P}\to\mathbb{R}$, where the both inputs are probability vectors. The first input of $G$ corresponds to the pre-query probability, i.e., the probability currently predicted by the model, while the second input corresponds to the post-query probability. For instance, if we query ``Class'' to $x$, and the label of $x$ is $c$. Then the pre-query probability vector is $p(x;\theta)$ and the post-query probability vector is $e_c$, where $e_c$ is the $c$th basis vector of length $C$. However, since the true class of $x$ is unknown prior to querying, we estimate the information gain by taking the expectation:
$$\text{Gain}(x;\text{Class},\theta) := E_{p(x;\theta)}\left[G(p(x;\theta)||e_c)\right] = \sum_{c=1}^C p_c(x;\theta)G(p(x;\theta)||e_c).$$
If we take $G$ to be the KL-divergence or the total variation,
$$G(p||r)=\sum_{l=1}^Cr_l\log \left(\frac{r_l}{p_l}\right) \quad\text{or}\quad G(p||r)=\sum_{l=1}^C\frac{1}{2}|p_l-r_l|,$$
respectively, then the entropy and the variance criteria can be recovered.

Let $P_{1:m},R_{1:m}\in\mathbb{P}^{m}$ be two matrices, each with rows representing probability vectors. For convenience, we refer to $P$ and $R$ as probability matrices. Define $G(P_{1:m}||R_{1:m})=\sum_{i=1}^mG(P_{i\cdot}||R_{i\cdot})$ where $P_{i\cdot}$ is the $i$th row of $P$. Let $\mathcal{P}(q_k,a_k,Q_k)$ denote the set of probabilities for which $\Pr(a_k|q_k)$ is 1 within the given family. Clearly, for ``Class'' question, $\mathcal{P}(x,c,``\text{Class}'') = \{e_c\}$. For ``All'' question, we have
\begin{align*}
	&\mathcal{P}(\{x_{1:m},c\},a=1,\text{All}) = \{(e_c^T,\dots,e_c^T)^T\}\\
	&\mathcal{P}(\{x_{1:m},c\},a=0,\text{All}) = \{P\in\mathbb{P}^{m}:\text{at least one }P_{ic}=0, \text{ for }i=1,\dots,m\}.
\end{align*}
For ``Any'' question, we have
\begin{align*}
	&\mathcal{P}(\{x_{1:m},c\},a=1,\text{Any}) = \{P\in\mathbb{P}^{m}:\text{at least one }P_{ic}=1, \text{ for }i=1,\dots,m\}\\
	&\mathcal{P}(\{x_{1:m},c\},a=0,\text{Any}) = \{P\in\mathbb{P}^{m}:\text{All }P_{ic}=0, \text{ for }i=1,\dots,m\}.
\end{align*}
By defining 
$$G(P||\mathcal{P})= \min_{R\in\mathcal{P}}G(P||R),$$
the information gain for the three types of questions can be expressed in a unified form as
$$\text{Gain}(q;Q_k,\theta) = E_a\left[G(p(q;\theta)||\mathcal{P}(q,a,Q_k))\right]=\sum_{a\in \mathcal{A}_k}\Pr(a|q;\theta)G(p(q;\theta)||\mathcal{P}(q,a,Q_k)).$$

The intuition behind the approach is to predict what the probability vector (matrix) will be after querying $Q_k$ to $q$. If the true answer is $a$, then the post-query probability of $a$ should be one. Hence, we need to find all probability vectors (or matrices) for which the probability of $a$ equals one; this set is denoted by $\mathcal{P}(q, a, Q_k)$. \ {Since there may be multiple probability vectors (or matrices) in $\mathcal{P}(q, a, Q_k)$, each with a different information gain, we conservatively select the one with the minimum gain to avoid overestimating the potential informativeness.} Lastly, since we have assumed the true answer is $a$--which is unknown before querying--we take the expectation of the information gain over all possible answers $a$, weighted by their estimated probabilities. Exhaustively searching all possible combinations to find the optimal realization $q$ is infeasible. Instead, the exchanging algorithm introduced in Subsection S3.2 of the supplementary material is applied to efficiently obtain an approximate optimum.

With the well-defined information gain for all types of questions, our active learning algorithm proceeds in two steps. Given the candidate set $\mathcal{D}_{can}^x\subset\mathcal{D}_{full}^x$, the first step determines the optimal realization $q_k^*$ for each $k=0,\dots, \tilde k$ by selecting points from $\mathcal{D}_{can}^x$ that maximize $\text{Gain}(q_k;Q_k,\theta(\mathcal{D}))$ for each $k=0,\dots, \tilde k $. The second step determines the querying question $k^*$ by randomly sampling from $k=0,\dots, \tilde k $ with probability $\Pr(k)\propto \text{Gain}^2(q_k^*;Q_k,\theta(\mathcal{D}))$. With the two steps, the question $Q_{k^*}$ with realization $q_{k^*}^*$ will be queried.

\subsection{Exploration and exploitation method}
\label{subsec:E and e}
When the model is insufficiently accurate, the discrepancies between the predicted probabilities and the true probabilities can be substantial, leading to a large gap between the estimated and true losses. In such cases, the active learning algorithm may waste the budget to focus on exploiting the uninformative area. To mitigate this risk, the model should expand its knowledge by exploring underrepresented regions of the sample space. To facilitate this, we introduce a distance function $dist$, whose specific form will be detailed at the end of the subsection. \ {Recall the definition of $\mathcal{D}_{full}^x$ and $\mathcal{D}=\{\mathcal{D}_0,\mathcal{D}_1,\dots,\mathcal{D}_{\tilde k}\}$. Similarly, define $\mathcal{D}^x=\cup_{k=0}^{\tilde k} \mathcal{D}_k^x$.} For any $x\in \mathcal{D}_{full}^x$, define its distance to $\mathcal{D}^x$ as 
$$dist(x,\mathcal{D}^x)=\min_{x'\in \mathcal{D}^x} dist(x,x').$$
A small value of $dist(x, \mathcal{D}^x)$ indicates that some information has already been acquired in the neighborhood of $x$. Therefore, to promote exploration, we should consider querying points $x$ for which $dist(x, \mathcal{D}^x)$ is not small.

Once the sample space has been sufficiently explored, the focus should gradually shift to exploitation, although determining the appropriate time to make this transition can be challenging. We propose a data-driven exploration and exploitation framework that leverages distance as a mechanism to gradually transition from exploration to exploitation. Given a sequence of distance thresholds $d_1> d_2>\cdots > d_S=0$ and a proportion threshold $\rho$, each active learning iteration screens out points that are close to $\mathcal{D}^x$, and treats the remaining points as the candidate set. This is achieved by constructing $\mathcal{D}_{can}^x(s)=\{x\in \mathcal{D}_{full}^x:dist(x,\mathcal{D}^x)>d_s\}$ for $s=1,\dots,S$. We then identify the smallest $s$ such that $|\mathcal{D}_{can}^x(s)| > \rho N$. In other words, we require that at least a proportion $\rho$ of the data remains available for selection. The procedure of the exploration and exploitation framework is summarized in Algorithm~\ref{alg: ee}.
\begin{algorithm}[ht]
  \SetAlgoLined
  \SetAlgoNlRelativeSize{0}
  \SetNlSty{textbf}{(}{)}
  \KwData{$\mathcal{D}_{full}^x$, $\mathcal{D}_{0}^x$, $\rho$, $S$, $q$ and distance function $dist$.}
  \KwResult{Candidate set $\mathcal{D}_{can}$.}
  Set $d_1=\text{quantile}(dist(x,x'),q)$ and $d_1>\cdots>d_S=0$ by arithmetic sequence.\;
Identify the minimum $s$ such that $|D_{can}^x(s)|>\rho |D_{full}^x|$and set $\mathcal{D}_{can} = D_{can}^x(s)$.

  \caption{The exploration and exploitation framework}
  \label{alg: ee}
\end{algorithm}

\begin{figure}[ht]
\begin{center}
\includegraphics[width=6in]{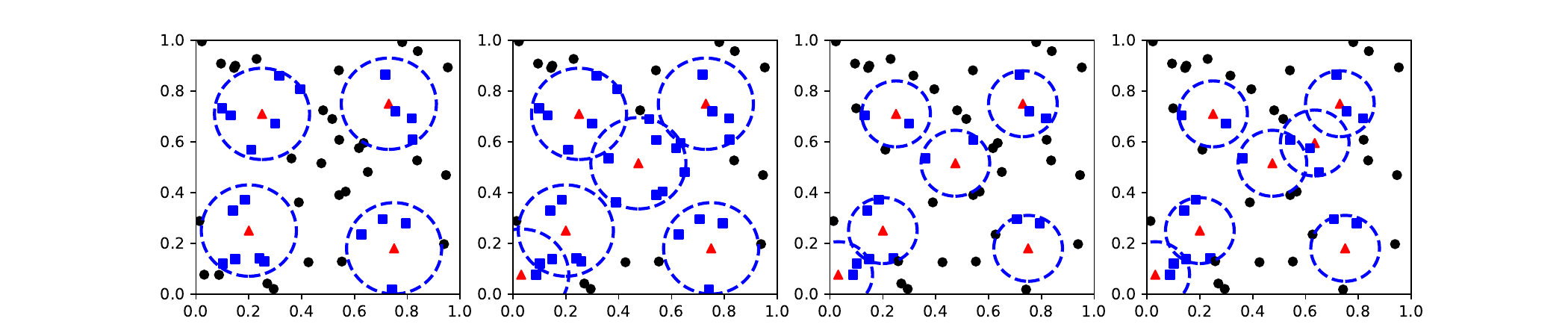}
\caption{Illustration of the exploration and exploitation framework. \label{fig:Eande}}
\end{center}
\end{figure}

Figure~\ref{fig:Eande} provides a visual example. The red triangles indicate the points in $\mathcal{D}^x$. In the first panel, the blue dashed circles have radius $d_1$, and the black dots denote candidate points in $\mathcal{D}_{can}^x(1)$. The blue squares are excluded as they are too close to $D^x$. The second panel displays that some black dots from the first panel are selected for querying. The blue dashed circles with radius $d_1$ are also shown in the second panel. However, these circles cover too many points, resulting in the size of $\mathcal{D}_{can}^x(1)$ being less than $\rho N$. Therefore, in the third panel, the radius is reduced to $d_2$ and the size of $\mathcal{D}_{can}^x(2)$ once again exceeds $\rho N$. In the fourth panel, the active learning algorithm continues querying a question within $\mathcal{D}_{can}^x(2)$. This exploration and exploitation framework is maintained throughout the active learning process until the budget is exhausted.

The distance thresholds are hard-to-decide parameters. We present a data-driven approach to select the distance thresholds. We first compute all pairwise distances within $\mathcal{D}_{full}^x$ and set the $5\%$ quantile as $d_1$. Setting $d_S=0$, we generate $d_2,\dots,d_S$ using an arithmetic sequence. In our simulation, we set $S=6$. The remaining hyper-parameter, $\rho$, controls the trade-off between exploration and exploitation. A larger $\rho$ places more emphasis on exploitation. We set $\rho = 0.25$ throughout the remainder of this paper. A detailed discussion of the hyper parameters is provided in Subsection S3.1 of the supplementary material.

The final component of the framework is to determine an appropriate distance metric. While Euclidean and Mahalanobis distances are viable options, they are unsupervised and sensitive to high dimensionality. Introducing a supervised distance metric is essential, as the input space may not yield a meaningful notion of distance, particularly in the context of image data. For a broad class of probabilistic models, the predicted probabilities are generated through an activation function $\sigma$,
\begin{equation}
\label{equ:softmax model}
\ {
	p_c(x;\theta)=\frac{\sigma\{h_c(x;\theta)\}}{\sum_{l=1}^C\sigma\{h_l(x;\theta)\}}.}
\end{equation}
Under such models, the logits $h(x;\theta) = (h_1(x;\theta), \dots, h_L(x;\theta))$ can be used to define a model-guided distance
$$dist(x,x';\theta)=\Vert h(x;\theta)-h(x';\theta)\Vert_2,$$
where $\Vert\cdot\Vert_2$ denotes the $L_2$-norm.

\subsection{Extension to batch active learning}
\label{subsec: batch AL}
The exploration and exploitation framework can also be adapted to batch active learning, where multiple questions are queried before updating the model. It operates in a simple yet effective manner by adding each queried question along with its corresponding answer into the training set, without refitting the model until the batch budget is exhausted. The extension to batch active learning is shown in Algorithm \ref{alg:Proposed AL}, along with the full procedure. For instance, one can simply set line (6) to proceed every $\tilde b$ queries, thereby implementing batch active learning with batch size $\tilde b$. Batch active learning may suffer from redundant information if the active learning criterion is applied greedily. Under the exploration and exploitation framework, each queried point generates an exclusive region that screens out nearby redundant points. If the exploration and exploitation framework is not employed, simply \ {set} $\mathcal{D}_{can}^x=\mathcal{D}_{full}^x-\mathcal{D}^x$ in line (3). To apply the exploration and exploitation framework within traditional active learning, it suffices to define a single question, ``Class'', and modify the gain function in line~(4) to the desired criterion.
\begin{algorithm}[ht]
  \SetAlgoLined
  \SetAlgoNlRelativeSize{0}
  \SetNlSty{textbf}{(}{)}
  \KwData{Budget $B$, proportion $\rho$, number of thresholds $S$, full pool $\mathcal{D}_{full}^x$, gain function $\text{Gain}$, current information $\mathcal{D}_k$ and cost $\text{cost}_k$ for $k = 0, 1, \dots, \tilde{k}$.}
  \KwResult{probabilistic model $p(\cdot,\;\hat\theta)$.}
Initialize model parameter $\hat\theta = \theta(\mathcal{D})$ and used budget $\text{budget}_{\text{used}} = 0$\;
  \While{$B \geq \text{budget}_{\text{used}} + \min\{\text{cost}_k\}$}{
  
      Generate candidate set $\mathcal{D}_{can}^x(s)$ via Algorithm~\ref{alg: ee}\;  
    Identify questions $Q_k$ satisfying $\text{budget}_{\text{used}} + \text{cost}_k \leq B$ and determine optimal realization $q_k^*$ by maximizing $\text{Gain}(q_k; Q_k, \hat\theta)$. Sample question index $k^*$ from all feasible $k$ with probability proportional to $\text{Gain}^2(q_k^*; Q_k, \hat\theta)$\;  
    Query realization $q_{k^*}^*$ and receive answer $a_{k^*}^*$. Update  $\mathcal{D}_{k^*} \leftarrow \mathcal{D}_{k^*} \cup \{(q_{k^*}^*, a_{k^*}^*)\}$ and $\text{budget}_{\text{used}} \leftarrow \text{budget}_{\text{used}} + \text{cost}_{k^*}$\;  
  Update the model parameter by $\hat\theta=\theta(\mathcal{D})$ (Optional).
  }
  \caption{New framework of active learning with exploration and exploitation}
  \label{alg:Proposed AL}
\end{algorithm}
%-----------------------------------------------------------------
\section{Theory}
\label{sec:Theory}
This section provides theoretical support for both the proposed active learning framework with multiple questions and the exploration and exploitation framework. \ {We aim to show that (1) the uncertainty upper bound rate of traditional active learning is better than that of random sampling, (2) the information gain of the three types of questions can be explicitly expressed under specific choices of the function $G$, (3) the proposed active learning framework in Algorithm~\ref{alg:Proposed AL} achieves a favorable uncertainty upper bound rate, and (4) the exploration and exploitation framework is theoretically validated.}

The information gain for the ``Class'' question also measures the uncertainty of a point $x$. Therefore, the uncertainty of a point $x$ is defined by
$$AL(x;\theta(\mathcal{D})) := \text{Gain}(x;\text{Class},\theta(\mathcal{D})) = \sum_{c=1}^C p_c(x;\theta(\mathcal{D}))G(p(x;\theta(\mathcal{D}))||\{e_c\}).$$
For simplicity, we sometimes write $AL(x;\mathcal{D})$ or $AL(x)$. Assume $\pi$ is the data distribution over $\mathcal{X}$ where $\mathcal{X}=[0,1]^d$. Let $S_b$ be the budget used after querying $b$ questions and $\mathcal{D}_{S_b}$ be the entire training set after $b$th querying. Assume the cost of ``Class'' question is 1. Thus, in traditional active learning $S_b=b$ always holds. Define $AL_{S_b} = E_{\pi}[AL(X;\mathcal{D}_{S_b})]$ and $c(x)=\operatorname{argmax}_c p_c(x;\mathcal{D}_{S_b})$. We list the following assumptions:
\begin{enumerate}[{A}1]
	\item $AL(x)\in[0,1]$.
	\item $\underline\pi:=\inf_{x\in\mathcal{X}}\pi(x)>0$ and $\bar\pi:=\sup_{x\in\mathcal{X}}\pi(x)<\infty$.
	\item If $x\in \mathcal{D}_0^x$, then $AL(x)\le \epsilon$ with some $\epsilon\ge 0$.
	\item \ {For any $x,x'\in\mathcal{X}$, we have $|AL(x)-AL(x')|\le K dist^\alpha(x,x')$.}
	\item For any $b$,
	\begin{align*}
		&\int_{\mathcal{X}} \mathbf{1}\left[\left\{AL(x|\mathcal{D}_{S_{b}})-\epsilon\right\}_+-\left\{AL(x|\mathcal{D}_{S_{b+1}})-\epsilon\right\}_+\right] \pi(x)dx\\
		\le& K_0 \left[\int_{\mathcal{X}}\left\{AL(x|\mathcal{D}_{S_{b}})-\epsilon\right\}_+\pi(x)dx\right]^t,
	\end{align*}
	with some constant $K_0$ and  $t>\alpha/d+1$.
	\item $\Pr(Y=c(X)|X)\ge p_{c(X)}(X;\theta(\mathcal{D}_{S_b}))-K'AL_{{S_b}}^{t'}$ where $t'>0$.
\end{enumerate}
Assumption A1 is standard, since most uncertainty measures are finite and can always be rescaled to $[0,1]$. Assumption A2 guarantees that the support of $\pi$ coincides with $\mathcal{X}$. Assumptions A3 and A4, which respectively assume low uncertainty on the training set and a Lipschitz continuity condition, are common in active learning literature such as \cite{sener2017active}, and can be reasonably satisfied by most neural networks. Assumption A5 states that only a small portion of points in $\mathcal{X}$ experience an increase in uncertainty as more information is acquired. Assumption A6 implies that when uncertainty is high, even a high predicted probability may not correspond to the true class; conversely, when uncertainty is low, a high predicted probability is more likely to be correct.

\subsection{Traditional active learning}
\label{subsec: theory traditional AL}
In traditional active learning, given the uncertainty measurement $AL(\cdot)$, the $(b+1)$th queried sample is $x_{b+1}^* = \operatorname{argmax}_{x\in\mathcal{D}_{full}^x}AL(x;\mathcal{D}_{S_b})$. Theorem \ref{thm: uncertainty upper bound of AL} reveals that the uncertainty upper bound in traditional active learning decays at the rate of $O(B^{-\alpha/d})$.
\begin{Theo}
\label{thm: uncertainty upper bound of AL}
	Under \ {assumptions A1 to A5}, suppose the full data $\mathcal{D}_{full}^x$ is independently and identically sampled from $\mathcal{X}$, and the initial training set $\mathcal{D}_0$ of size $n$ is uniformly randomly drawn from $\mathcal{D}_{full}$. Then, there exist constants $n_0$ and $N_0$ such that for all $n \ge n_0$ and $N \ge N_0$, with probability at least $1-K_2 (N\log N)^{-1}-K_1 (n\log n)^{-1}$, the uncertainty satisfies the upper bound
	$$AL_{S_B}\le \left(K_3B+K_4 (n/\log n)\right)^{-\alpha/d}+\epsilon,$$
	where $K_1,K_2,K_3$ and $K_4$ are constants independent of $n,N,B$.
\end{Theo}
The randomness in the theorem arises from the sampling of the full pool and the initial training set. Once these are fixed, the subsequent active learning process becomes deterministic. According to the proof of Theorem~\ref{thm: uncertainty upper bound of AL}, it suffices to set $n_0=O(1)$ and require $B\le O(N_0/\log N_0)$. Such a requirement is reasonable, as active learning inherently assumes a large pool size and a relatively small labeling budget.  Theorem~\ref{thm: uncertainty upper bound of AL} further reveals that traditional active learning achieves a faster uncertainty upper bound decay rate compared to random sampling. This can be readily verified by setting $n \leftarrow n + B$ and $B \leftarrow 0$, which corresponds to allocating all budget to random sampling.

\begin{figure}[ht]
\begin{center}
\includegraphics[width=6in]{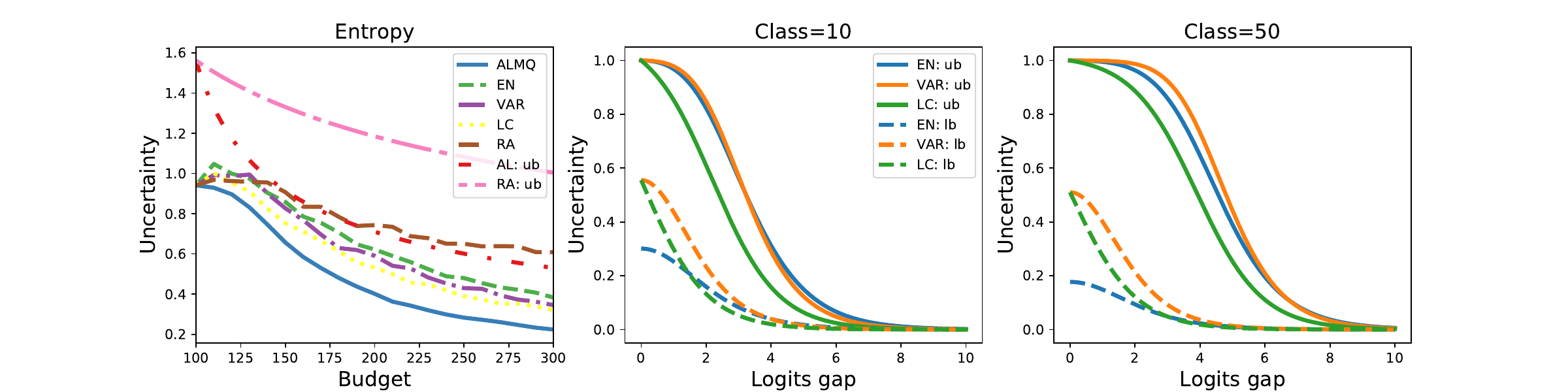}
\end{center}
\caption{The left panel displays the uncertainty curves on the MNIST dataset, and the middle and right panels are uncertainty bounds based on logits gap derived in Theorem~\ref{thm: Uncertainty bound}.\label{fig: uncertainty_eevalie}}
\end{figure}

The left panel of Figure~\ref{fig: uncertainty_eevalie} shows the uncertainty curves of different active learning methods on the MNIST dataset \citep{lecun1998mnist}. The $d/\alpha$, $K_3$, and $K_4$ in Theorem \ref{thm: uncertainty upper bound of AL} are set to $2$, $1/12$, and $1/10$, respectively. A detailed introduction of the active learning methods is deferred to Section~\ref{sec:Simulation}. ``AL: ub'' and ``RA: ub'' denote the uncertainty upper bound for active learning and random, respectively, as established in Theorem \ref{thm: uncertainty upper bound of AL}. The proposed active learning method exhibits lower uncertainty, particularly under the entropy criterion, and all active learning methods outperform random sampling in terms of uncertainty. Similar conclusions hold for the variance and least-confidence criteria, and thus their results are omitted from the figure.

\subsection{Active learning with multiple questions}
We now turn to the information gain criterion tailored for the proposed multi-question framework. Recall that for a question $Q_k$ with realization $q$ and answer $a$, we identify the family of all possible post-query probabilities $\mathcal{P}(q,a,Q_k)$, and conservatively select the probability vector (matrix) within the family that minimizes the information gain. Since the number of elements in the family can be infinite, it is essential to derive an explicit expression for the information gain given $(q,a)$. We focus on the ``All'' and ``Any'' questions, as the explicit expression of ``Class'' has already been established in $AL(\cdot)$.
\begin{Theo}
\label{thm: expression of gain}
Suppose $G(p||r) = \sum_{c=1}^Cg(p_c,r_c)$ and $q=\{x_{1:m},c\}$. Then
\begin{enumerate}
	\item If $g(a,b) = \phi(b-a)$ where $\phi(\cdot)$ is convex, then
	\begin{align*}
		\text{Gain}&(q;ALL,\theta) = \left\{\prod_{i=1}^m p_c(x_{i};\theta) \right\}\sum_{i=1}^m \left\{\phi(1-p_c(x_i;\theta))+\sum_{l\neq c}\phi(-p_l(x_i;\theta))\right\}\\
		&+\left\{1-\prod_{i=1}^m p_c(x_{i};\theta) \right\}\min_{i=1,\dots,m} \left\{(C-1)\phi\left(\frac{p_c(x_i)}{C-1}\right)+\phi(-p_c(x_i;\theta))\right\}  \ {\text{, and}}\\
		\text{Gain}&(q;ANY,\theta) = \left\{\prod_{i=1}^m (1-p_c(x_{i};\theta)) \right\}\sum_{i=1}^m \left\{(C-1)\phi\left(\frac{p_c(x_i)}{C-1}\right)+\phi(-p_c(x_i;\theta))\right\}\\
		&+\left\{1-\prod_{i=1}^m (1-p_c(x_{i};\theta)) \right\}\min_{i=1,\dots,m} \left\{\phi(1-p_c(x_i;\theta))+\sum_{l\neq c}\phi(-p_l(x_i;\theta))\right\}.
	\end{align*}
	\item If $g(a,b)=b\phi(a/b)$ where $\phi(\cdot)$ is convex and $g(a,b)\to g_0$ as $b\to 0$, then
	\begin{align*}
		\text{Gain}&(q;ALL,\theta) = \left\{\prod_{i=1}^m p_c(x_{i};\theta) \right\}\sum_{i=1}^m \left\{\phi(p_c(x_i;\theta))+(C-1)g_0\right\}\\
		&+\left\{1-\prod_{i=1}^m p_c(x_{i};\theta) \right\}\min_{i=1,\dots,m}\left\{\phi(1-p_c(x_i))+g_0\right\}\ {\text{, and}}\\
		\text{Gain}&(q;ANY,\theta) = \left\{\prod_{i=1}^m (1/p_c(x_{i};\theta)) \right\}\sum_{i=1}^m \left\{\phi(1-p_c(x_i))+g_0\right\}\\
		&+\left\{1-\prod_{i=1}^m (1-p_c(x_{i};\theta)) \right\}\min_{i=1,\dots,m} \left\{\phi(p_c(x_i;\theta))+(C-1)g_0\right\}.
	\end{align*}
\end{enumerate}	
\end{Theo}

For example, the total variation $g(a,b)=1/2|a-b|$ and KL-divergence $g(a,b)=-b\log(a/b)$ correspond to the two representative choices, respectively.

With the explicit expressions of information gain available for each question, we now investigate the uncertainty rate under the proposed framework. The setting involving more than one point can be analytically challenging. We primarily focus on the scenario where two types of questions are available: ``what is class of $x$?'' and ``is $x$ from class $c$'', with respective costs $1$ and $c_1\le 1$. We refer to the second question as ``Is''. Hereafter, we adopt the total variation gain function $g(a,b)=1/2|a-b|$, which is also employed in the simulations in Section~\ref{sec:Simulation} and real world applications in Section~\ref{sec: real data}. The following lemma indicates that, under the ``Is'' question, the optimal choice of class $c$ for a given $x$ is the most probable class.
\begin{Lem}
\label{lem: optimal c}
\ {For a given $x$}, the optimal choice of $c$ for question ``Is $x$ from class $c$?'' under $g(a,b)=|a-b|/2$ is $\operatorname{argmax}_l p_l(x;\theta)$.
\end{Lem}
With the discussions of Theorem \ref{thm: expression of gain} and Lemma \ref{lem: optimal c}, we can derive the uncertainty rate of the proposed framework, with the similar technique as in Theorem \ref{thm: uncertainty upper bound of AL}.
\begin{Theo}
\label{thm: ALMQ uncertainty bound}
Under {assumptions A1 to A6} with $\epsilon=0$ and all other assumptions in Theorem \ref{thm: uncertainty upper bound of AL}, let the information gain be total variation. Assume two types of questions are available: ``Class'' with cost $1$ and ``Is'' with cost $c_1\le 1$. Denote by $AL_{last}$ the final uncertainty after completing the active learning procedure. Then, there exist constants $n_0'$ and $N_0'$ such that for $n\ge n_0'$ and $N\ge N_0'$, with probability at least $1-K_2' (N\log N)^{-1}-K_1' (n\log n)^{-1}-1/B$,
	 $$E[AL_{last}|\mathcal{D}_{S_0}]\le \left(K_3'\lfloor \max(\tilde B,B)\rfloor+K_4'(n/\log n)\right)^{-\alpha/d},$$
where $L=\frac{4+C^2c_1}{4+C^2c_1^2}$, $K_1',K_2',K_3'$ and $K_4'$ are constants independent of $n,N,B$, and
	 $$\tilde B = \frac{1}{2L^2 c_1^2}\left\{2B c_1 L+\frac{(1-c_1)^2}{2}\log B - \sqrt{2c_1L(1-c_1)^2B\log B+\frac{(1-c_1)^4}{4}(\log B)^2}\right\}.$$
\end{Theo}
In the theorem, the uncertainty arises primarily from two sources: the random sampling of the full data pool and initial samples, and the squared-based question selection strategy described in line (4) of Algorithm~\ref{alg:Proposed AL}. The expectation in the theorem is taken over the squared-based sampling procedure. When $c_1 = O(1)$, the uncertainty rate of the proposed framework matches that of traditional active learning. This is reasonable since the ``Is'' question can never yield more information than the ``Class'' question. Even though the ``Is'' question may have a lower cost, as long as the cost is $O(1)$, the total number of queries made by the proposed framework remains on the same order as in traditional active learning. Therefore, achieving the same uncertainty rate is expected. Nevertheless, when $c$ is small, the proposed framework achieves better uncertainty rate. 

\subsection{Validation of the exploration and exploitation framework}
Many classification models take the form given in Equation~(\ref{equ:softmax model}). The activation function $\sigma:\mathbb{R}\to\mathbb{R}^+$ is increasing and differentiable almost everywhere. A common choice is the ``softmax'' function, where $\sigma(\cdot)=\exp(\cdot)$. The proposed exploration and exploitation framework filters out data points close to the training set using the distance between logits. A natural question arises: do points close to the training set necessarily exhibit low uncertainty? Even if such points remain in the candidate set, their low uncertainty means they are unlikely to be queried. To address this, Theorem~\ref{thm: Uncertainty bound} and the middle and left panels of Figure~\ref{fig: uncertainty_eevalie} demonstrate that uncertainty is sensitive to the gap in logits. As long as points are not too close to the training set, their uncertainty can be significantly large. This is further confirmed by simulations showing substantial performance gains when incorporating the exploration and exploitation framework. Define the logits gap as $\delta(x) = h_{c(x)}(x) - \max_{c \neq c(x)} h_c(x)$. We seek to bound uncertainty in terms of $\delta(x)$ and the largest logit $h_{c(x)}(x)$. We focus on three common uncertainty criteria: entropy, variance, and least-confidence.
\begin{Theo}
\label{thm: Uncertainty bound}
Consider the entropy, variance, and least-confidence criteria evaluated at point $x$. For probabilities generated via a positive, increasing, and almost everywhere differentiable activation function $\sigma$, these three uncertainty measures admit tight upper bounds and near-tight lower bounds expressed in terms of the logits gap $\delta(x)$ and the largest logit $h_{c(x)}(x)$. Moreover, if $\sigma(\cdot) \propto \exp(\cdot)$, then these bounds depend solely on the logits gap $\delta(x)$.
\end{Theo}
For clarity, the explicit formulas are omitted here. The tight upper and near-tight lower bounds can be verified in the proof. The upper and lower bounds generated by the ``softmax'' activation with 10 and 50 classes are illustrated in the middle and right panels of Figure~\ref{fig: uncertainty_eevalie}. For ease of comparison, all uncertainty criteria are rescaled to the interval $[0,1]$. The upper bounds on uncertainty can be small even for moderate logits gaps, e.g., around 10. Although a training point may exhibit a moderate logits gap, a neighboring point can have a smaller gap, potentially leading to higher uncertainty. The proposed exploration and exploitation framework effectively screens out such points.

%-----------------------------------------------------------------
\section{Simulation}
\label{sec:Simulation}
The section presents simulations of the proposed active learning method along with the exploration and exploitation framework, evaluated on various models and datasets. Several baseline methods are compared against our proposed multi-question active learning method (ALMQ). The first group of baselines consists of common uncertainty-based methods, including entropy (EN), variance (VAR), and least confidence (LC) criteria \citep{shannon1948mathematical,budd2021survey}. The second baseline is a typical Bayesian method, known as Bayesian Active Learning by Disagreement (BALD) \citep{houlsby2011bayesian}. The third baseline, Batch Active Learning by Diverse Gradient Embeddings (BADGE) \citep{ash2019deep}, is a representative-based method that incorporates expected model change and is widely regarded as state-of-the-art. The final baseline is random sampling (RA). Our ALMQ method considers six questions: ``Class'', ``All'' for $m=1,2,3$ and ``Any'' for $m=2,3$. Since the ``All'' and ``Any'' questions coincide when $m=1$, the ``Any'' question for $m=1$ is excluded.

In this section, the MNIST and HAR datasets have already been pre-divided into training and testing sets. For the other dataset, we randomly split it into training and testing set with ratio 8:2. We consider the training set as the full pool $\mathcal{D}_{full}$, and evaluate all metrics on the testing set, which remains unseen by the model during training. This follows the standard protocol for active learning evaluation. In addition to active learning methods, we also train a model on the entire training set to establish an upper bound on the achievable performance for each dataset.

A common phenomenon in neural network models is over-confidence. This occurs because, with limited training data, the model is poorly calibrated, resulting in overly confident predictions that can be incorrect. Such over-confidence can undermine the reliability of our proposed expected information gain for various questions. To address this issue, we incorporate MCdrop \citep{gal2016dropout}, a method that only requires adding dropout layers to the neural network. Originally, MCdrop was proposed to enable efficient approximate Bayesian sampling in BALD-based methods. We utilize MCdrop to generate multiple sampled probability matrices for the data. For the ``All'' and ``Any'' questions, we select the sampled probability matrices whose estimated probabilities of the answers ``Yes'' and ``No'' are closest to the $40\%$ quantile, respectively. A detailed comparison of different MCdrop selection quantile choices and the effect of omitting MCdrop is provided in Subsection S2.1 of the supplementary material. The results indicate that the choice of quantile is robust as long as it is near $40\%$, and that employing MCdrop significantly improves performance compared to not using it.

For clarity, only active learning methods incorporating the exploration and exploitation framework are shown, except for the BADGE method, which is a representative-based criterion with its own way to balance exploration and exploitation. It should be emphasized that the performance of each method has been substantially improved compared to its original baseline, thanks to the exploration and exploitation framework. Comparisons between methods with and without the exploration and exploitation framework are presented only for the MNIST dataset, while results for other datasets are given in Subsection S3.1 of the supplementary material. Additionally, further simulations on hyper parameters are provided in Section S2 of the supplementary material.

\subsection{Feature data}
\label{subsec: ANN}
Two datasets from the UC Irvine machine learning repository are used in the subsection. The first dataset, Handwriting, consists of pre-processed handwritten digits from 0 to 9 \citep{handwriting} and is used to train a logistic regression model. The second dataset, HAR, involves human activity recognition using smartphone sensor data \citep{HAR} and is modeled by a shallow neural network. The detailed information, including initial data size, batch active learning size, total budget and the cost of each question, is shown in Subsection S3.3 of supplementary material.

Figure \ref{fig: hand_HAR} shows the accuracy curves for different methods. The $95\%$ confidence intervals for the mean accuracy are also shown. Except for BADGE, all other methods incorporate the exploration and exploitation framework. With the exploration and exploitation framework, traditional methods as well as the Bayesian method achieve comparable performance to, or even surpass, the state-of-the-art method BADGE. Our proposed ALMQ method outperforms all other methods by a substantial margin. On the HAR dataset, the Bayesian method BALD performs worse than random sampling. This is because the exploration and exploitation framework greatly improves random sampling, while BALD may overly focus on parameter uncertainty rather than accuracy. In summary, the exploration and exploitation framework enhances traditional active learning methods to achieve performance comparable to or better than the state-of-the-art method BADGE, while our ALMQ method significantly outperforms all.
\begin{figure}[ht]
\begin{center}
\includegraphics[width=6in]{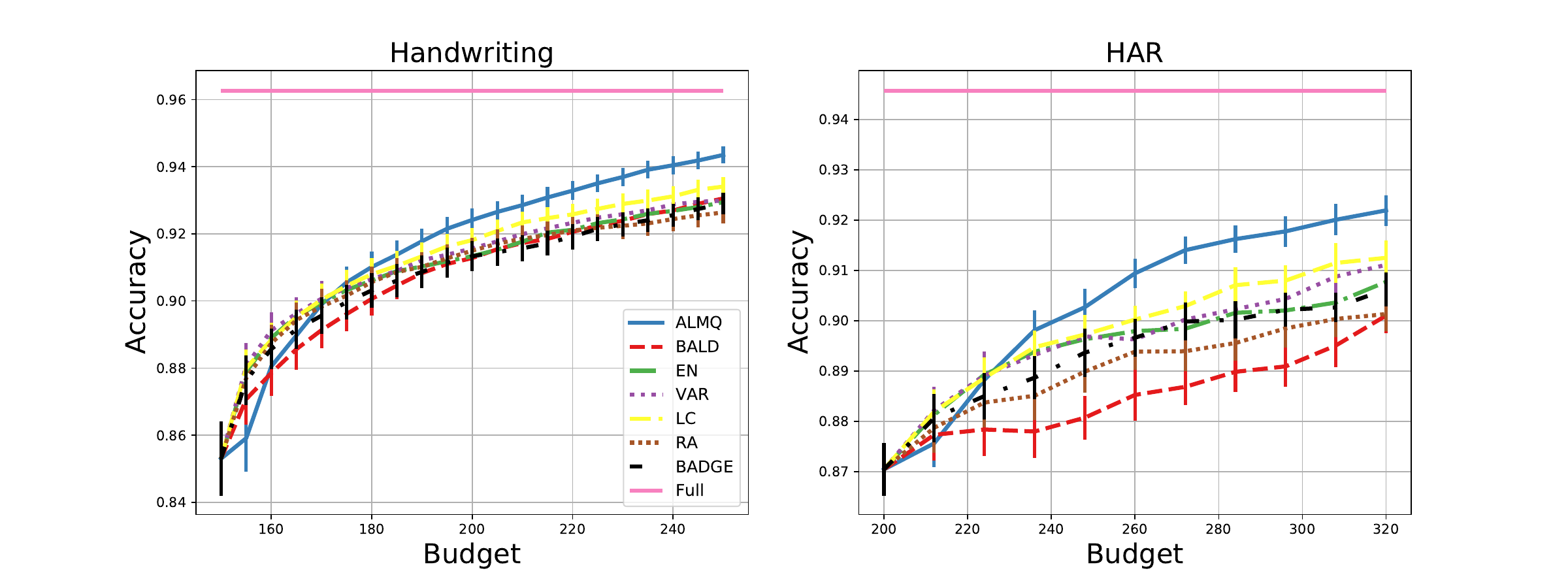}
\end{center}
\caption{Accuracy curves estimated on the testing data for the Handwriting and HAR dataset\label{fig: hand_HAR}}
\end{figure}

\subsection{Image data}
\label{subsec: CNN}
When dealing with high-dimensional and non-linear image data, convolutional neural networks (CNNs) are typically employed. The widely used $(28\times 28)$-dimensional MNIST dataset \citep{lecun1998mnist} is used in this subsection. 

\begin{figure}[ht]
\begin{center}
\includegraphics[width=6in]{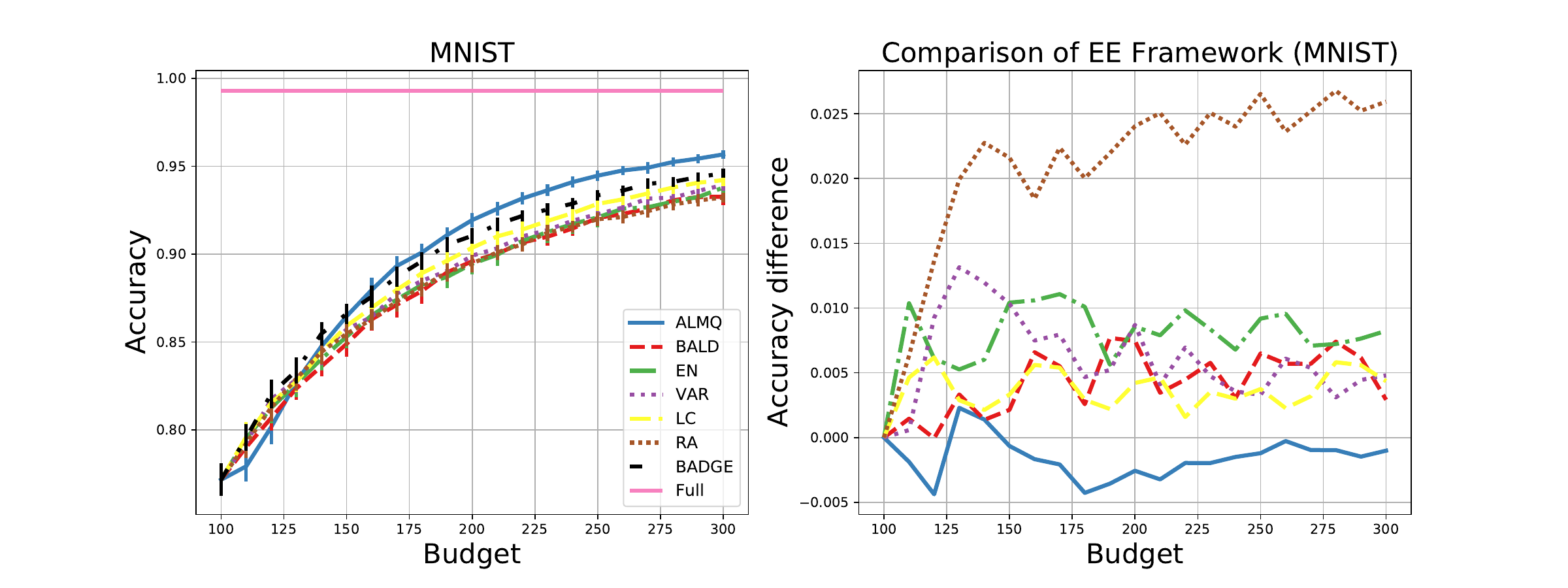}
\end{center}
\caption{Accuracy curves estimated on the testing data for the MNIST dataset (left penal) and the different between accuracies with or without the exploration and exploitation framework. \label{fig: MNIST_ee}}
\end{figure}

The left panel of Figure~\ref{fig: MNIST_ee} presents the accuracy curves. The conclusion is consistent with that in Subsection~\ref{subsec: ANN}. With the help of the exploration and exploitation framework, random sampling achieves nearly the same accuracy as several traditional active learning methods, while our proposed ALMQ method consistently outperforms all others. The right panel of Figure~\ref{fig: MNIST_ee} illustrates the difference in accuracy curves between methods with and without the exploration and exploitation framework. For the ALMQ method, the inclusion of the exploration and exploitation framework does not yield a significant difference. This is because the ALMQ method is capable of exploring the space through multi-point queries. Nevertheless, traditional methods, the Bayesian approach, and random sampling all benefit from the framework, particularly during the early stages of active learning. Random sampling experiences the greatest improvement, explaining why the RA methods in Figures~\ref{fig: hand_HAR} and \ref{fig: MNIST_ee} perform comparably to other methods. This also highlights that active learning methods significantly outperform naive random sampling when the framework is not applied.

%-----------------------------------------------------------------
\section{Real World Dataset}
\label{sec: real data}
Artificial neural networks and CNNs are typically employed for relatively simple datasets. For real world image datasets, deep neural networks are often used to capture complex dependencies among inputs. In this subsection, two real world datasets from Kaggle are used to illustrate the effectiveness of the ALMQ method in both daily life and medical domains. The first dataset is the Animals-10 dataset \citep{corrado_animals10}. The ResNet50 model \citep{he2016deep}, with several initial layers frozen, is applied to this dataset. The second dataset is the Brain Tumor MRI dataset \citep{nickparvar2021brain}, which includes images of three tumor types as well as normal MRI scans. The ResNet18 model is employed for this dataset. Transfer learning is applied only to the Animals dataset, as the pre-trained ResNet parameters are based primarily on everyday image datasets such as ImageNet \citep{deng2009imagenet}. It is inappropriate to directly transfer these pre-trained parameters to a distinct domain such as the medical field. Samples are randomly split into training and testing sets as in Section~\ref{sec:Simulation}, and all images are resized to $3\times 224\times 224$ in both datasets.

\begin{figure}[ht]
\begin{center}
\includegraphics[width=6in]{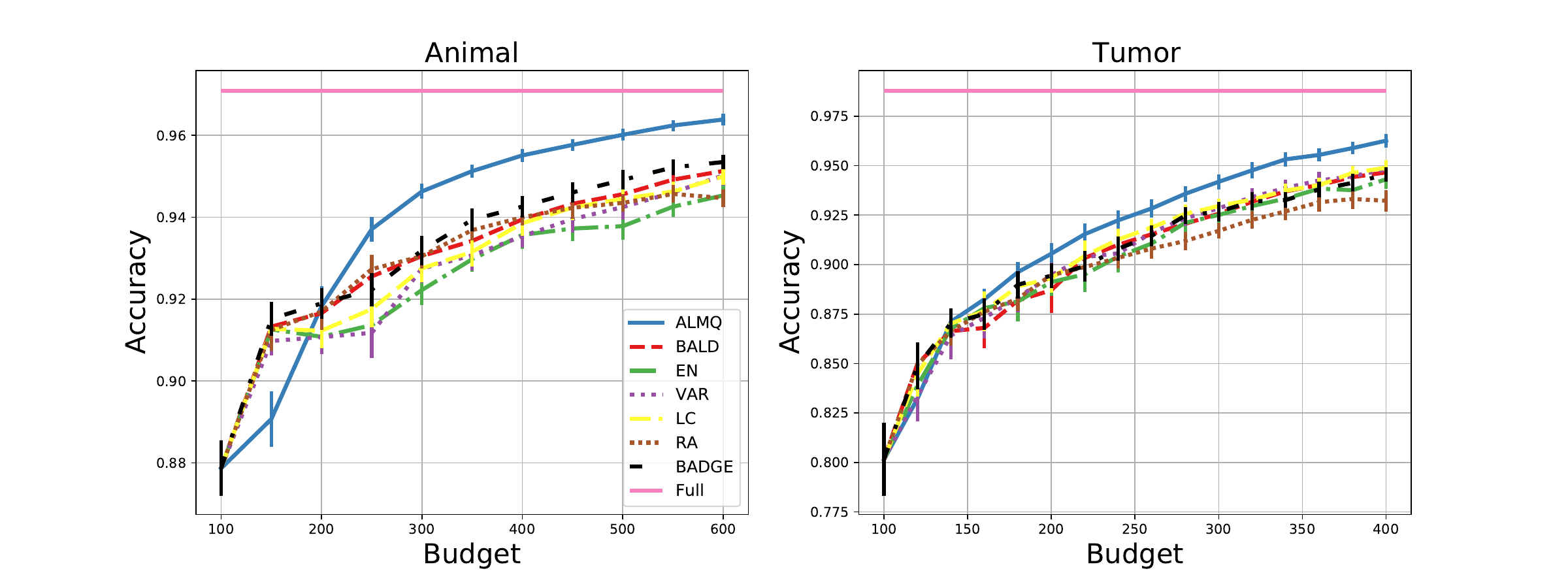}
\end{center}
\caption{Accuracy curves estimated on the testing data for the Animal and Tumor datasets. \label{fig: Animal_Tumor}}
\end{figure}

Analogous to the previous subsections, the accuracy curves of all methods are shown in Figure~\ref{fig: Animal_Tumor}. The ALMQ method consistently outperforms all other methods. In the Animals dataset, the ALMQ method with only 600 budget achieves nearly the same performance as the full dataset containing over 20000 samples. Although ALMQ does not exhibit strong performance in the first batch of active learning, it rapidly surpasses all other methods in the subsequent batches. The initial underperformance of ALMQ is again because of model over-confidence. During the initial stages of active learning, the model may encounter surprising answers containing only partial information. In the Tumor dataset, ALMQ performs comparably to other methods in the first two batches and improves rapidly thereafter. The accuracy gap between ALMQ and the full-data model is small, despite the latter using over 5000 samples.

In summary, the proposed ALMQ framework exhibits strong performance and consistently outperforms all other methods across the five datasets. The exploration and exploitation framework yields substantial improvements to traditional active learning methods. Across Figures~\ref{fig: hand_HAR}, \ref{fig: MNIST_ee}, and \ref{fig: Animal_Tumor}, the state-of-the-art method, BADGE, indeed performs well. Traditional active learning methods, when equipped with the exploration and exploitation framework, can match or even surpass the performance of BADGE. Aside from the ALMQ method, no single criterion exhibits consistently optimal performance. We recommend the VAR and LC methods due to their robust performance across all settings. The BALD and EN methods show unsatisfactory results in the HAR and Animal datasets, respectively. As discussed in Section~\ref{sec:method}, under the ``Class'' question, the VAR criterion is equivalent to the total variation information gain. This equivalence motivates our use of total variation information gain in simulations and theoretical developments in Subsection~\ref{subsec:AL}.

%-----------------------------------------------------------------
\section{Concluding Remarks}
\label{sec:Concluding Remarks}
In this paper, we address the classification active learning problems from a novel perspective on how the data are labeled. The three main contributions are summarized as follows. First, we propose a method to effectively integrate full and partial information into model building by using probabilities as a crucial bridge to connect partial information. Second, we introduce information gain tailored for ALMQ framework. The proposed ALMQ method considers multiple questions and selects among them based on conservative information gain and associated cost. Third, we propose an exploration and exploitation framework that can be embedded into various active learning criteria by screening out points with redundant information via a data-driven approach. In addition, we propose a model-guided distance metric that is applicable to a wide range of models. The model-guided distance is updated dynamically throughout the active learning process. Simulation results show that the exploration and exploitation framework significantly enhances traditional methods, enabling them to match or exceed the performance of state-of-the-art approaches. Moreover, ALMQ consistently outperforms all other methods across various models.

The cost of the ``All'' and ``Any'' questions is typically smaller than that of the ``Class'' question in the paper. This reflects a practical setting, especially for highly parallel human annotation. In some domains, such as ophthalmology, this parallel assumption does not hold because careful inspection of each image is required. However, even when parallel processing is not possible, if the number of classes $C$ is large, the cost of ``All'' and ``Any'' questions remains relatively low. In our experiments, we set the cost of each question to a reasonable level, reflecting these practical considerations. If annotation costs are high in practice, additional strategies, such as probability clipping, may be incorporated into the question sampling procedure.

This paper focuses on the setting of probabilistic models. The applicability of multi-question active learning in other settings, such as metric learning, remains unexplored. A major challenge is how to quantify partial information. A naive solution uses Gaussian assumptions to link distances with probabilities. Nevertheless, the Gaussian assumption may not hold in many scenarios, and developing more general approaches remains an open problem for future research. Another prerequisite in this paper is that the number of classes is known and fixed. This assumption has been previously discussed in related literature. For instance, \cite{sun2016online} and \cite{mohamad2018active} highlight that in streaming data, class distributions may evolve, with old classes vanishing and new ones emerging. As a result, models need to adaptively identify and handle such dynamic class scenarios. A further assumption is that experts serve as perfect oracles, which can be violated in practice due to issues such as misdiagnosis. Finally, the number of classes considered is relatively small compared to the sample size. An interesting direction for future work is to extend the framework to settings with a large number of classes \citep{deng2009imagenet}. 

\section*{Disclosure statement}
The authors report there are no competing interests to declare.

\bibliographystyle{agsm}

\bibliography{Bibliography-ALMQ}
\end{document}